\documentclass[twoside]{article}

\usepackage{PRIMEarxiv}

\usepackage[utf8]{inputenc} % allow utf-8 input
\usepackage[T1]{fontenc}    % use 8-bit T1 fonts
\usepackage{hyperref}       % hyperlinks
\usepackage{url}            % simple URL typesetting
\usepackage{booktabs}       % professional-quality tables
\usepackage{amsfonts}       % blackboard math symbols
\usepackage{nicefrac}       % compact symbols for 1/2, etc.
\usepackage{microtype}      % microtypography
\usepackage{lipsum}
\usepackage{fancyhdr}       % header
\usepackage{graphicx}       % graphics
\graphicspath{{media/}}     % organize your images and other figures under media/ folder

\usepackage{amsmath}
\usepackage[colorinlistoftodos]{todonotes}
    
\usepackage{subcaption}

\usepackage{amssymb} % for mathbb
\usepackage{tcolorbox} % for color table
\usepackage{makecell}
\PassOptionsToPackage{ruled,vlined,linesnumbered}{algorithm2e}
\usepackage{algorithm2e}
\usepackage{booktabs} 
\usepackage{microtype}
\usepackage{graphicx}
\usepackage{booktabs} % for professional tables
\usepackage{wrapfig}
\title{SCoOP: Semantic Consistent Opinion Pooling for Uncertainty Quantification in Multiple Vision-Language Model Systems
}

\author{Chung-En Johnny Yu\\
    University of West Florida\\
    Pensacola, Florida, USA\\
    \texttt{cy31@students.uwf.edu} \\
\And
    Brian Jalaian\\
    University of West Florida\\
    Pensacola, Florida, USA\\
    \texttt{bjalaian@uwf.edu} \\
\AND
    Nathaniel D. Bastian\\
    United States Military Academy\\
    West Point, New York, USA\\
    \texttt{nathaniel.bastian@westpoint.edu} \\
}

\begin{document}
\maketitle

\begin{abstract}
Combining multiple Vision–Language Models (VLMs) can enhance multimodal reasoning and robustness, but aggregating heterogeneous models' outputs amplifies uncertainty and increases the risk of hallucinations. 
We propose \textbf{SCoOP} (\emph{Semantic-Consistent Opinion Pooling}), a \emph{training-free} uncertainty quantification (UQ) framework for multi-VLM systems through uncertainty-weighted linear opinion pooling. 
The core idea is to treat each VLM as a probabilistic ``expert," sample multiple outputs, map them to a unified space, aggregate their opinions, and produce a system-level uncertainty score.
Unlike prior UQ methods designed for single models, SCoOP explicitly measures collective, system-level uncertainty across multiple VLMs, enabling effective hallucination detection and abstention for highly uncertain samples. 
On ScienceQA, SCoOP achieves an \textbf{AUROC of 0.866} for hallucination detection, outperforming baselines (\textbf{0.732-0.757}) by approximately \textbf{10-13\%}. 
For abstention, it attains an \textbf{AURAC of 0.907}, exceeding baselines (\textbf{0.818-0.840}) by \textbf{7-9\%}. 
Despite these gains, SCoOP introduces only microsecond-level aggregation overhead relative to the baselines, which is trivial compared to typical VLM inference time (on the order of seconds). 
These results demonstrate that SCoOP provides an efficient and principled mechanism for uncertainty-aware aggregation, advancing the reliability of multimodal AI systems.
Our code is publicly available at \href{https://github.com/chungenyu6/SCoOP}{https://github.com/chungenyu6/SCoOP}.
\end{abstract}
% keywords can be removed
\section{Introduction}
\label{sec:intro}
\vspace{-5pt}

%%%%%%%%%%%%%%%%%%%%%%%%%%%%%%%%%%%%%%%%%%%%%%%%%%%%%%%%%%%%%%%%%%%%%%%%%%%%%%%%%%%%%%%%%%%%%%%% 
\begin{wrapfigure}{r}{0.4\columnwidth}
    \vspace{-20pt}
    \centering
    \includegraphics[width=0.85\linewidth]{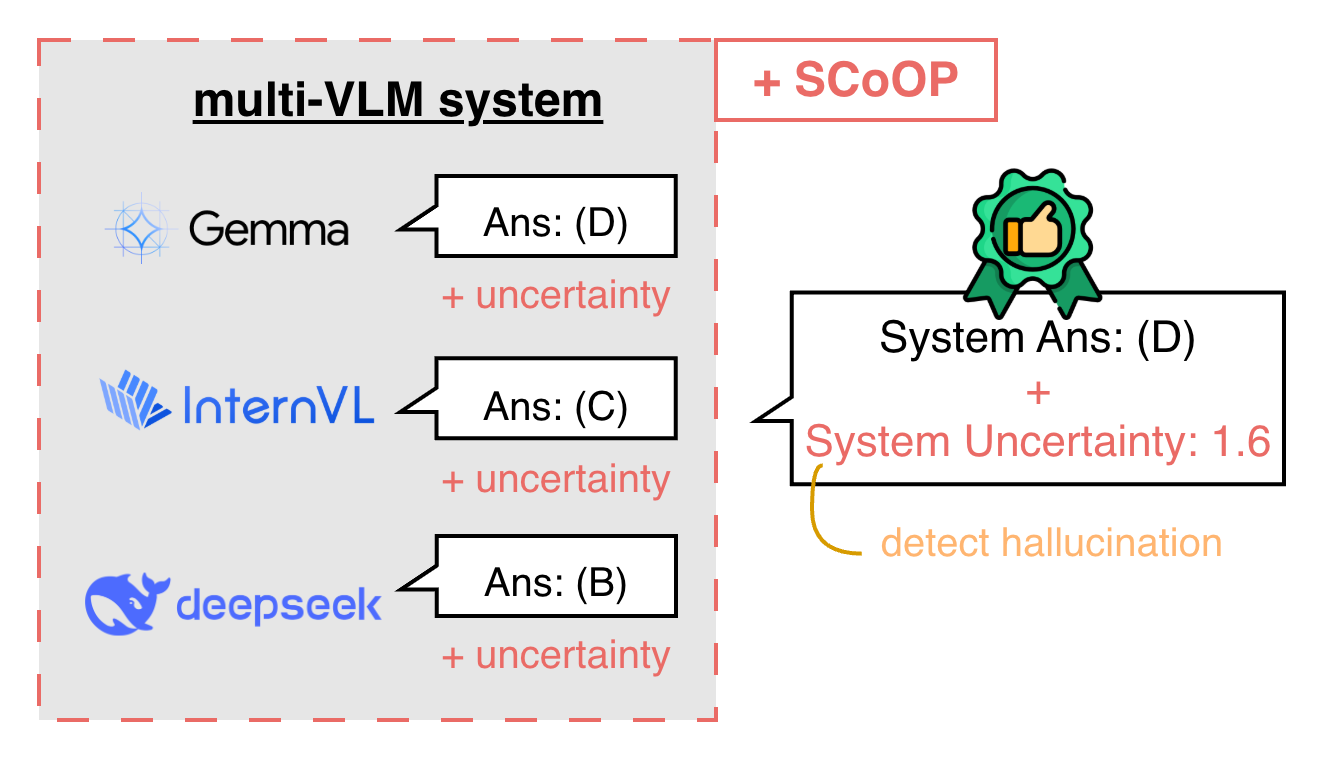}
    \vspace{-8pt}
    \caption{
        \footnotesize{
        \textbf{Overview of SCoOP increasing the reliability of a multi-VLM system.}}
    }
    \label{fig:scoop-overview}
    \vspace{-10pt}
\end{wrapfigure}
%%%%%%%%%%%%%%%%%%%%%%%%%%%%%%%%%%%%%%%%%%%%%%%%%%%%%%%%%%%%%%%%%%%%%%%%%%%%%%%%%%%%%%%%%%%%%%%%

%By Dr. J to remove the focus on Multi Agent
Despite the recent remarkable advancements in Vision-Language Models (VLMs), hallucination remains an open challenge, rendering them unreliable for real-world deployment in safety-critical applications such as medical diagnosis and multi-robot collaboration \cite{li2025survey, chen2024unveiling, liu2025uncertainty}.
Hallucination occurs when a VLM outputs information that is not supported by the input image, such as non-existent objects, incorrect attributes, or false relationships.
Detecting VLM hallucinations requires not only semantic understanding of natural language but also alignment with visual evidence \cite{liu2024survey}.
Uncertainty quantification (UQ) has emerged as a crucial approach for detecting hallucinations by revealing the uncertainty inherent in model responses, which is essential for risk control \cite{zhang2024vl, shorinwa2025survey}.

Beyond individual models, recent research has increasingly explored \emph{multi-VLM aggregation}, which combines the outputs of heterogeneous VLMs to enhance multimodal reasoning and robustness \cite{liu2025advances, xie2024large, chen2025harnessing}.
Multi-VLM systems share structural similarities with early-stage multi-agent systems, in which multiple models contribute to a shared decision without explicit coordination.
As highlighted by \cite{lu2024merge}, uncertainty-guided aggregation across multiple VLMs yields notable gains in Visual Question Answering (VQA) tasks.
However, the same work also cautions that improper aggregation can amplify individual model errors and propagate hallucinations, ultimately leading to unreliable system behavior.
This observation naturally extends the role of UQ from single-model settings to multi-VLM systems, where uncertainty must be quantified not only at the model level but also at the system level \cite{shorinwa2025survey}.

Most existing UQ methods depend on additional training or labeled data, as exemplified by Bayesian neural networks and conformal prediction \cite{shorinwa2025survey}. 
Even for a single generative model, these requirements introduce substantial computational overhead and engineering complexity. 
When scaled to multi-VLM systems, where multiple models must be jointly considered, such approaches become increasingly impractical due to training costs and data dependencies.
% Among various UQ methodologies, consistency-based UQ methods are particularly attractive because they require no additional training or labeled data, making them practical for resource-limited deployment \cite{liu2025uncertainty}.
% Their practicality makes them strong candidates for extending UQ beyond individual VLMs, motivating us to explore how such UQ methods can be effectively adapted to systems that have multiple heterogeneous VLMs.
To the best of our knowledge, there is no existing UQ method explicitly for detecting hallucinations in  multi-VLM systems.
To address this gap, we propose a training-free framework that quantifies system-level uncertainty to enhance reliability.

\noindent \textbf{Contributions.} 
We introduce \textbf{SCoOP} (\emph{Semantic-Consistent Opinion Pooling for Uncertainty Quantification}), a training-free framework for quantifying uncertainty in multi-VLM systems, as illustrated in Figure~\ref{fig:scoop-overview}. 
(1) We propose an uncertainty-weighted linear opinion pooling mechanism that aggregates multiple VLMs' probabilistic outputs over a unified class space, yielding a system-level uncertainty estimate. 
(2) \textbf{SCoOP achieves an AUROC of 0.866} for hallucination detection, outperforming baselines (0.732-0.757) by approximately \textbf{10-13\%}, and an \textbf{AURAC of 0.907} for abstention, surpassing baselines (0.818–0.840) by \textbf{7-9\%}, illustrating how robust uncertainty enables the model to abstain from high-uncertainty predictions to maintain high accuracy.
(3) Despite these gains, SCoOP introduces only microsecond-level aggregation overhead relative to the baselines, which is negligible compared to the typical VLM inference time (on the order of seconds), demonstrating practical deployment efficiency.

% The remainder of this paper is organized as follows. 
% Section~\ref{sec:related} reviews prior work on UQ for large language and vision-language models, highlighting the gap in existing methods for multi-VLM systems. 
% Section~\ref{sec:method} introduces the proposed \textbf{SCoOP} framework and its uncertainty-weighted opinion pooling formulation. 
% Section~\ref{sec:exp} presents experimental results on ScienceQA, demonstrating the effectiveness and efficiency of SCoOP for hallucination detection and abstention. 
% Finally, Section~\ref{sec:conclusion} concludes the paper and outlines future research directions.

\vspace{-5pt}
\section{Related Work}
\label{sec:related}
\vspace{-5pt}
\textbf{UQ for LLMs.}
With the rapid progress of Large Language Models (LLMs), UQ has been developed to detect hallucinations and improve model's reliability in safety-critical applications, such as robotic collaboration \cite{liu2025uncertainty}.
For practical implementation, we specifically explore approaches that are training-free and do not require access to model weights or labeled data.
Existing approaches fall into three broad categories:
(1) \textit{Token-level UQ} \cite{fadeeva2024fact, ling2024uncertainty}, which quantifies uncertainty from the model's output token probability distributions that could detect hallucinations.
However, \cite{wang2025predictive, finlayson2024logits} reveal that commercial model providers might restrict access to token probabilities to reduce potential security risks.
(2) \textit{Self-verbalized UQ} \cite{lin2022teaching, stengel2024lacie}, where models articulate their own confidence. 
While intuitive, this approach often demands extra data or fine-tuning for calibration \cite{shorinwa2025survey}.
(3) \textit{Consistency-based UQ} \cite{lin2023generating, farquhar2024detecting, xiong2023can}, which quantifies uncertainty by measuring semantic similarity across multiple generations. 
This category is particularly attractive due to its training-free nature and without the need for token probability access. 
Moreover, unlike conformal prediction \cite{su2024api}, consistency-based UQ does not require additional calibration data, making it more practical for real-world deployment.

\textbf{UQ for VLMs.}
In contrast, UQ for hallucination detection in VLMs remains relatively underexplored \cite{liu2025uncertainty, li2024reference, xiao2025detecting}.
Most existing works use uncertainty to support training \cite{ji2023map},  decoding \cite{fang2024uncertainty, li2025mitigating}, or reasoning \cite{zhi2025seeing}.
The most notable exception is VL-Uncertainty \cite{zhang2024vl}, a consistency-based approach specifically detecting VLM hallucinations by sampling VLM for multiple generations with perturbed multimodal inputs.
The same work adapts Semantic Entropy \cite{farquhar2024detecting} from LLM to VLM settings for comparison by sampling VLM with the same input.
Moving beyond the UQ for individual VLM, our proposed UQ framework, SCoOP, aggregates multiple VLMs' uncertainties and responses for detecting hallucinations in multi-VLM systems.

\textbf{UQ for Multi-VLM Systems.}
Recent studies \cite{pan2025multiagent, duan2025enhancing, chen2025harnessing} underscore producing reliable aggregated outputs in language model-based multi-agent systems is still an open challenge, where they imply leveraging UQ is a promising strategy for tackling this problem.
In practice, multi-VLM systems can be viewed as an intermediate stage between single models and autonomous multi-agent systems, where multiple specialized models jointly contribute to a shared output.
As such, challenges studied in multi-agent reliability naturally arise in multi-VLM aggregation settings.
To detect hallucinations for revealing the system outputs' reliability, one should quantify the uncertainty of the entire system.
% To our knowledge, there are limited studies \cite{qu2025uq, zhang2025consensus} that leverage uncertainty to aggregate multiple VLMs.
A recent study, named UQ-Merge \cite{qu2025uq}, proposes an uncertainty-guided process that aggregates outputs from multiple VLMs. 
However, it does not quantify the overall uncertainty of the multi-VLM system, which makes its aggregated output's reliability remain unknown.
Another work, Consensus Entropy \cite{zhang2025consensus}, ensembles VLMs' responses weighted by uncertainties, resulting in superior accuracy compared to individual VLMs.
However, the authors do not evaluate how the uncertainty estimates impact system-level hallucination detection or abstention, and their evaluation is limited to the OCR task domain.
In contrast, our work focuses on the effectiveness of system-level uncertainty as a hallucination indicator across a wide range of VQA tasks, including OCR, visual grounding, reasoning, etc.
The lack of an effective method for quantifying the uncertainty of the entire multi-VLM system represents a critical gap in the literature.
This motivates us to propose a novel UQ framework, SCoOP, that effectively quantifies the uncertainty for the entire multi-VLM system, addressing the needs outlined in Sec.~\ref{sec:intro}.

\begin{figure*}
    \centering
    \includegraphics[width=0.9\linewidth]{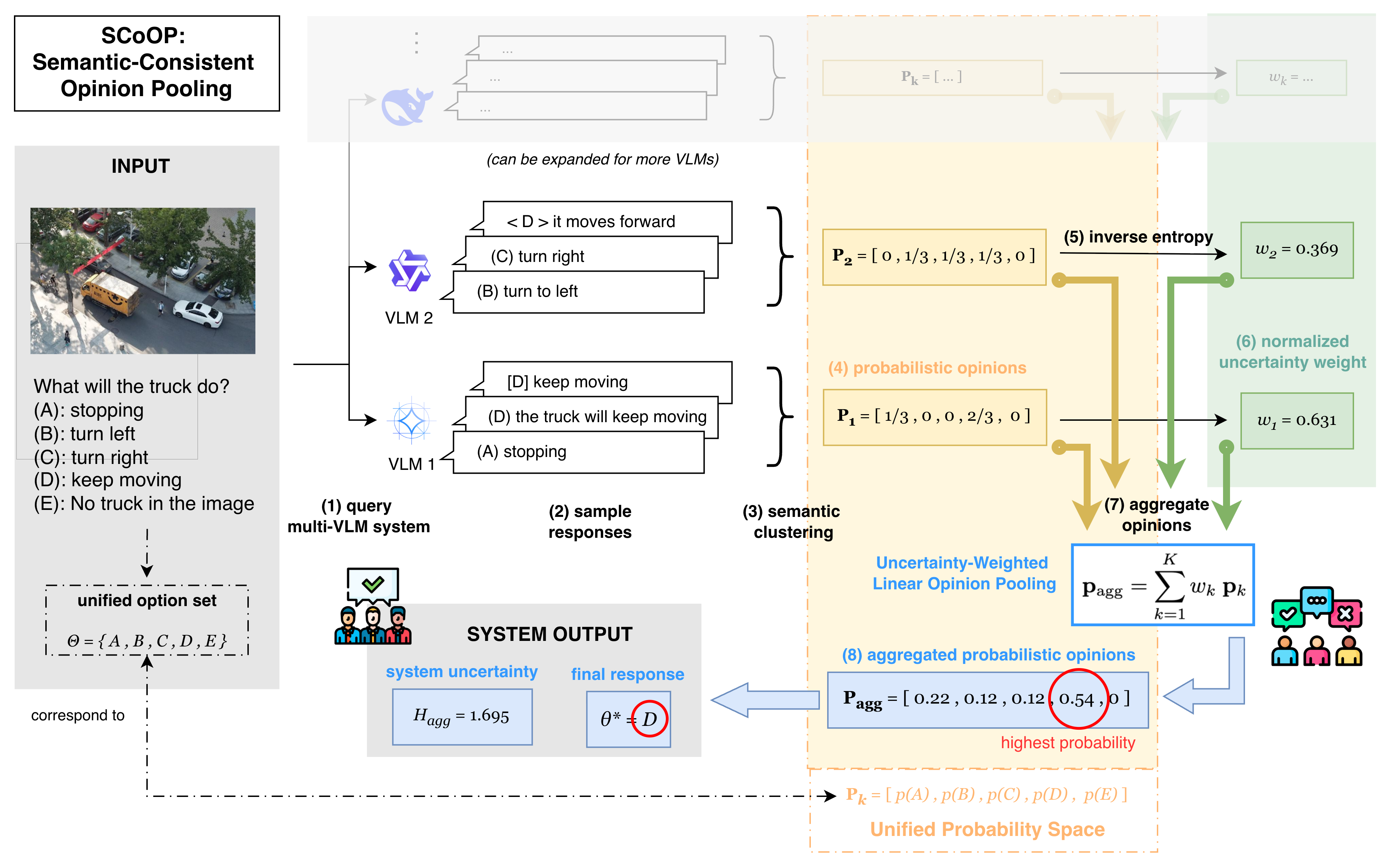}
    \caption{
        \footnotesize{
        \textbf{SCoOP (Semantic-Consistent Opinion Pooling) workflow.} 
        For each VLM $M_k$, we sample $N$ responses, map them to the unified option set $\Theta$, and form a probability vector $\mathbf{p}_k$. 
        Each model's uncertainty weight $w_k$ is calculated from its Shannon entropy.
        We then apply weighted linear opinion pooling $\mathbf{p}_{\text {agg }}=\sum_{k=1}^K w_k \mathbf{p}_k$ to obtain the system distribution, which can obtain the final response $\theta^*$ and the system-level uncertainty $H_{agg}$. 
        It unifies heterogeneous VLMs' outputs and quantifies uncertainty of the multi-VLM system.
        }
    }
    \label{fig:temp}
    \vspace{-15pt}
\end{figure*}

\vspace{-5pt}

\section{SCoOP Framework}
\label{sec:method}
\vspace{-5pt}

\textbf{Overview.}
An effective method is still lacking for quantifying the uncertainty of a multi-VLM system while aggregating their responses without amplifying individual errors or hallucinations.
Our proposed framework, \textbf{SCoOP} (\textit{Semantic Consistent Opinion Pooling for Uncertainty Quantification}), addresses this by introducing an \emph{uncertainty-weighted linear opinion pooling} strategy that quantifies both individual-model and system-level uncertainty.
The overall workflow is illustrated in Fig.~\ref{fig:temp}, and the corresponding pseudo-code is provided in Appendix~\ref{sec:appendix_a}.

SCoOP views each VLM as a probabilistic \textit{expert} that provides opinions over discrete answers.
Let $\mathcal{M}=\{M_1,\ldots,M_K\}$ denote a system of $K$ VLMs.
Each model $M_k$ receives the same multiple-choice question (MCQ) input $x=(I,q)$, consisting of an image $I$ and a text question $q$, and generates $N$ responses $Y_k=\{y_k^1,\ldots,y_k^N\}$.
Each response $y_k^n$ corresponds to a MCQ option $\theta^j$, where the options are denoted as a set $\Theta = \{\theta^{1}, \ldots, \theta^{J}\}$.
For each model $M_k$ given the input $x$, the probability of choosing option $\theta^j$ is the proportion of sampled responses that correspond to that option:
\begin{equation}
\label{eq:p_k}
  p_k(\theta^j \mid x)
   = \frac{1}{N}\sum_{n=1}^{N}
     \mathbf{1}\!\big[y_k^n
     \text{ corresponds to } \theta^j\big],  
\end{equation}
where the indicator $\mathbf{1}[\cdot] \in \{0,1\}$ returns $1$ if the condition holds true and $0$ otherwise. 
For notational brevity in subsequent sections, we suppress the dependence on $x$ and write $\mathbf{p}_{k}$ when the input is clear from context.
This results in the probability vector $\mathbf{p}_k=\big[p_k(\theta^1),\ldots,p_k(\theta^{J})\big]$.

This representation aligns naturally with the classical \textit{opinion pooling} formulation \cite{stone1961opinion,carvalho2012consensual}, where expert $k$ provides a probability distribution $\mathbf{p}_k$ over a shared hypothesis space $\Theta$ that contains mutually exclusive classes.
In linear opinion pooling (LOP), the aggregated distribution is
$$
\mathbf{p}_{\text{LOP}}
  = \sum_{k=1}^{K} w_k\,\mathbf{p}_{k}, \
  \text{ where} \ w_k \ge 0,\ 
  \sum_{k=1}^{K} w_k=1.
$$
Traditional LOP assumes identical class spaces for all experts, which does not hold for heterogeneous VLMs.
SCoOP extends this formulation by 
(i) unifying heterogeneous classes into a unified space $\Theta$ for estimating each model's probabilistic opinions, then
(ii) defining model weights $w_k$ according to each model's entropy-based uncertainty, thereby producing a system-level probability estimate, and
(iii) eventually quantifying the multi-VLM system's uncertainty for hallucination detection.

\vspace{-3pt}

\subsection{Single-VLM's Probabilistic Opinions}
\vspace{-3pt}

\textbf{Sampling Responses to Corresponding Options.}
SCoOP starts with sampling responses from each VLM to obtain its probabilistic opinions on the MCQ options.
For each $M_k$ given the same input $x$, it generates $N$ responses $Y_k=\{y_k^1,\ldots,y_k^N\}$ at temperature $T$.
The temperature controls how deterministic or stochastic the model's response is.
The model is formatted to output a MCQ option $\theta^j$ by using a zero-shot prompt.
See the prompt details in Appendix ~\ref{sec:appendix-prompts}.
For example in Fig.~\ref{fig:temp}, model $M_{1}$ samples three responses, which yields $Y_{1}$ = \{\textit{(A) stopping}, \textit{(D) the truck will keep moving}, \textit{[D] keep moving}\} and $M_{2}$ obtains $Y_{2}$ = \{\textit{(B) turn to left}, \textit{(C) turn right}, \textit{$<$ D $>$ it moves forward}\}.
Despite formatting the model to output MCQ options, raw responses may include extraneous text, such as explanations or malformed outputs. 
To address this, we follow the semantic clustering technique from \cite{farquhar2024detecting}, by applying a regular expression filter to extract the corresponding MCQ option $\theta^j$ from each response $y_{k}^{n}$.
In MCQ settings, each option corresponds to a distinct semantic meaning.
Thus, a simple regular expression filter suffices for semantic clustering. 
This yields the corresponding MCQ options while ensuring robustness against semantic variations in responses.
For example in Fig.~\ref{fig:temp}, model $M_{1}$'s responses ``$\textit{(A) stopping}$" correspond to option $(A)$; whereas ``$\textit{(D) the truck will keep moving}$" and ``$\textit{[D] keep moving}$" correspond to option $(D)$.

\textbf{Formulating Single Model's Opinions.}
As explained in Equation~\ref{eq:p_k}, we obtain the empirical probability $p_k(\theta^j \mid x)$ of $M_k$ choosing an option $\theta^j$ given the input $x$.
We then obtain the full probability distribution over all the options:
$\mathbf{p}_{k} = \big[ p_k(\theta^{1}), \ldots, p_k(\theta^{J}) \big].$
This represents $M_k$'s probabilistic opinions over all MCQ options $\Theta$.
For any options absent from the responses, their corresponding probabilities are set to zero.
This ensures that heterogeneous VLMs can be aggregated within a unified probability space.
As illustrated in Fig.~\ref{fig:temp}, model $M_1$ has one response corresponding to option $(A)$ and two responses to option $(D)$.
With $N=3$ responses, $M_{1}$ yields the probability distribution $\mathbf{p}_{1} = \big[ \frac{1}{3}, 0, 0, \frac{2}{3}, 0 \big]$.
This indicates that $M_{1}$'s opinions favor option $(D)$, with $66\%$ of its generated responses belonging to the option $(D)$, vice versa for other options.

\vspace{-3pt}

\subsection{Multi-VLM Opinions Aggregation}
% \johnny{add equation number for important equations: w_k, p_agg, theta*, H_agg}
\vspace{-3pt}

\textbf{Uncertainty-Weighted Linear Opinion Pooling.}  
We extend linear opinion pooling to incorporate each VLM's uncertainty as its weight.
We follow \cite{farquhar2024detecting} to quantify the uncertainty of each $M_k$ by computing the Shannon entropy with the preceding probabilities:
$
H_{k} = -\sum_{j=1}^{J} p_{k}(\theta^{j}) \log p_{k}(\theta^{j}).
$
Although prior studies \cite{padhi2025calibrating, shorinwa2025survey} have shown that entropy does not perfectly align with accuracy, a positive correlation between high entropy and hallucinations has been repeatedly observed \cite{farquhar2024detecting, liu2025uncertainty}.

Heuristically, each model's confidence score can be defined as the inverse of its entropy: $\mathrm{conf}_{k} = \frac{1}{H_{k}}$.
This inverse-entropy weighting provides a simple yet interpretable confidence measure that scales monotonically with model consistency.
The confidence scores are normalized across all $K$ models to obtain aggregation weights for each model $M_{k}$:
\begin{equation}
w_{k} = \frac{\mathrm{conf}_{k}}{\sum_{k'=1}^{K} \mathrm{conf}_{k'}} \ ,
\end{equation}
which satisfies the assumptions in opinion pooling where $w_k \ge 0$ and $\sum_{k=1}^{K} w_k=1.$
This uncertainty-weighting scheme penalizes VLMs with high-entropy (potentially hallucinated), allowing SCoOP to automatically down-weight less reliable models in the system.
With the uncertainty-weighted linear opinion pooling, the aggregated probability distribution can be expressed as:
\begin{equation}
\mathbf{p}_{\mathrm{agg}} 
= \sum_{k=1}^{K} w_{k} \ \mathbf{p}_{k} \ 
= \big[ p_{\mathrm{agg}}(\theta^{1}), \ldots, p_{\mathrm{agg}}(\theta^{J}) \big].
\end{equation}
It represents the overall multi-VLM system's aggregated opinions over all the MCQ options $\Theta$.

\textbf{Selecting Final Response.}  
The final system prediction $\theta^*$ is chosen as the option with the highest aggregated probability:
\begin{equation}
\theta^{*}=\arg \max_{\theta^{j} \in \Theta} \ p_{\text{agg}}(\theta^{j}).
\end{equation}
In Fig. \ref{fig:temp}, the final response is the option $(D)$, which yields the highest probability of $0.54$ among the options.
In cases where multiple options share the same maximum probability, SCoOP resolves ties by selecting the class favored by the VLM with the lowest individual entropy $H_k$.  
If the tie persists, a fixed ordering over the options is used for reproducibility.  

\textbf{System Uncertainty.}
Similar to quantifying the individual model's uncertainty, the system uncertainty can be obtained from the entropy of the aggregated probability distribution:
\begin{equation}
H_{\mathrm{agg}}=-\sum_{j=1}^{J} p_{\mathrm{agg}}\big(\theta^{j}\big) \log p_{\mathrm{agg}}\big(\theta^{j}\big).
\end{equation}
A lower $H_{\text{agg}}$ indicates a stronger consensus among the VLMs, whereas a higher value signals collective uncertainty or disagreement.
Directly using raw entropy for evaluation can be misleading because the number of possible answer options $J$ differs across input questions, making the resulting entropy values incomparable.  
To ensure scale-invariant comparison, we normalize the entropy by dividing it by its theoretical maximum:
$H_{\mathrm{agg}}^{\mathrm{norm}}=\frac{H_{\mathrm{agg}}}{\log _{2} J} \  \in[0,1].$

\vspace{-3pt}
\vspace{-5pt}
\section{Experiment}
\label{sec:exp}
\vspace{-3pt}
\subsection{Experiment Setup}
\vspace{-3pt}

\textbf{Baselines.}
\label{sec:exp_setup}
As mentioned in Sec.~\ref{sec:related}, existing works cannot be applied as baselines due to the lack of system-level uncertainty. 
Thus, we compare SCoOP against two heuristic, training-free uncertainty aggregation methods that output the uncertainty score of their predictions: \textit{Naive Selection} and \textit{Majority Voting}.
In general, they adopt Semantic Entropy \cite{farquhar2024detecting} to measure the uncertainty of the individual model, then aggregate responses and uncertainty.
(1) 
We follow an empirical practice from recent uncertainty-aware multi-agent systems studies \cite{zhi2025seeing, han2024towards} by adopting a heuristic ``Naive Selection" scheme: the system selects the model with the lowest uncertainty score among multiple models without aggregation in the probability space.
This aims to elicit the effectiveness of SCoOP's aggregation, which considers all the models in the system.
(2)
Following a common aggregation practice in LLM ensemble \cite{li2024more, ai2025beyond}, we adopt majority voting to aggregate the VLMs' uncertainties.
Each VLM casts one vote for its top-1 probable option after sampling responses.
% The final prediction is the class receiving the most votes, with ties broken by selecting the top aligned probability among tied classes.
The majority vote probability can be estimated with the empirical proportion of models voting for each option.
The uncertainty is estimated by the entropy of the majority vote probability, reflecting the level of agreement among models.
Full mathematical definitions and illustrations of the baselines are provided in Appendix~\ref{sec:appendix_a}.

\textbf{Evaluation Datasets.}
We evaluate the proposed uncertainty aggregation method on three widely recognized multimodal benchmarks that collectively capture diverse aspects of visual reasoning, perception, and real-world understanding.
\textit{ScienceQA}~\cite{lu2022learn} is a multimodal reasoning benchmark designed to assess a model's ability to integrate textual and visual information in scientific contexts. 
% It contains multiple-choice questions drawn from elementary to high school science curricula, covering subjects such as physics, earth science, and biology. 
% Each question may include accompanying text passages, diagrams, or images that require cross-modal interpretation. 
We use 952 representative samples from ScienceQA for evaluation.
\textit{MMMU}~\cite{yue2024mmmu} examines the generalization and college-level knowledge reasoning capabilities of VLMs across a wide range of academic and expert domains. 
% It features college-level multiple-choice questions spanning more than 20 disciplines, including medicine, engineering, and mathematics, often accompanied by tables or technical diagrams. 
% The benchmark tests not only factual recall but also visual reasoning and domain-specific comprehension. 
We evaluate our methods on a subset of 825 questions from MMMU.
\textit{MMBench}~\cite{liu2024mmbench} focuses on assessing real-world visual understanding, perception, and reasoning. 
% It encompasses a diverse set of tasks that reflect practical multimodal applications, such as scene understanding, human-object interaction, and chart interpretation. 
To ensure consistency with prior works, we evaluate our models on 950 multiple-choice samples drawn from MMBench.

\textbf{Metrics.}
We evaluate the quality of uncertainty estimates of the system with the following metrics:
(1) 
To measure how effectively uncertainty distinguishes hallucinated (incorrect) responses, we follow prior works \cite{farquhar2024detecting, xiong2023can} and report the Area Under the ROC Curve (AUROC).
AUROC values range from 0 to 1, where 1.0 indicates perfect hallucination detection and 0.5 corresponds to random guessing.
(2) 
We further employ the Area Under the Rejection Accuracy Curve (AURAC) \cite{xue2025verify, farquhar2024detecting} to evaluate how task accuracy improves as increasingly uncertain samples are rejected through abstention.
A higher AURAC reflects more informative and better-ranked uncertainty estimates.
(3)
We measure the trade-off between efficiency and performance with the common latency metric applied in multiple model systems: End-to-End Latency (E2E-Latency).
It is defined as the elapsed time from sending a request to receiving the final system uncertainty.
For each VLM system, we sum each model's inference latency and aggregation in a sequential manner, and report the 50th percentile (@p50) to capture the average-case scenarios.

\textbf{Models.}
To evaluate the effectiveness of SCoOP in varying multi-VLM settings, we employ sixteen state-of-the-art (SOTA) open-weighted VLMs spanning five distinct models: \textit{LLaVA-v1.6}, \textit{Gemma-3}, \textit{InternVL3}, \textit{DeepSeek-VL2}, and \textit{Qwen2.5-VL}. 
These models are categorized by parameter scale into four tiers: 
\textit{Small} (2–4B), \textit{Medium} (12-16B), \textit{Large} (27–38B), and \textit{Extra-Large} (72–78B). 
% For fair comparison, each tested multi-VLM system contains models of similar parameter scale. 
Within each scale, we systematically construct all possible cross-model combinations for $K \in \{2,3,4,5\}$ models, and denote a system containing $K$ models as a \textit{$K$-VLM (system)}.
% For instance, in the \textit{Small} scale category, six distinct combinations of 2-VLM systems can be formed across four models.
In total, this study evaluates 52 unique multi-VLM system combinations across parameter scales and $K\in\{2,3,4,5\}$ configurations.
A complete list of models and combination counts is provided in Appendix~\ref{app:exp_detail} Table~\ref{tab:vlm_combo}.
We report the average performance across all combinations of the same scale.
Further details regarding the experimental setup and computational resources are provided in Appendix~\ref{app:exp_detail}.
%%%%%%%%%%%%%%%%%%%%%%%%%%%%%%%%%%%%%%%%%%%%%%%%%%%%%%%%%%%%%%%%%%%%%%%%
%%%%%%%%%%%%%%%%%%%%%%%%%%%%%%%%%%%%%%%%%%%%%%%%%%%%%%%%%%%%%%%%%%%%%%%%
\vspace{-3pt}

\subsection{Result Analysis}
\vspace{-3pt}

%%%%%%%%%%%%%%%%%%%%%%%%%%%%%%%%%%%%%%%%%%%%%%%%%%%%%%%%%%%%%%%%%%%%%%%%%%%%%%%%%%%%%%%%%%%%%%%%
\begin{wrapfigure}{r}{0.45\columnwidth}
    \vspace{-15pt}
    \centering
    \includegraphics[width=1.0\linewidth]{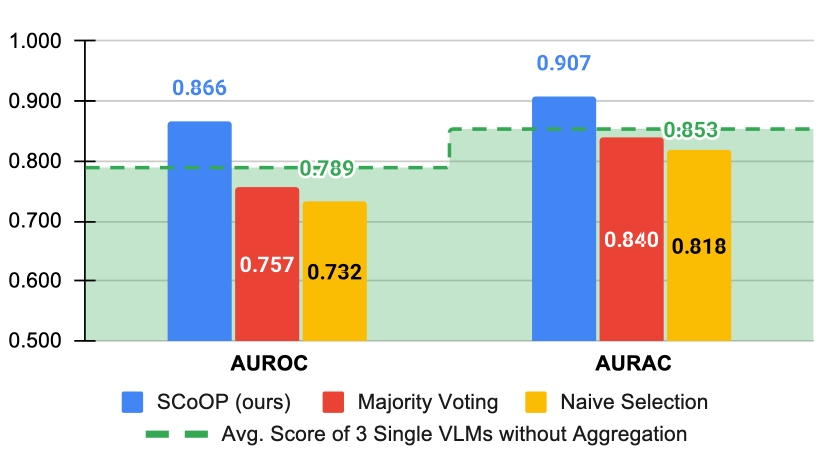}
    \caption{
        \footnotesize{
        \textbf{Hallucination detection (AUROC) and abstention (AURAC) performance on ScienceQA with 3-VLM systems of extra-large-parameter models.}
        Higher values indicate better uncertainty quality and impacts. The green dashed line marks the average AUROC/AURAC of the three individual VLMs evaluated separately (without aggregation).
        }
    }
    \label{fig:result_1}
    \vspace{-10pt}
\end{wrapfigure}
%%%%%%%%%%%%%%%%%%%%%%%%%%%%%%%%%%%%%%%%%%%%%%%%%%%%%%%%%%%%%%%%%%%%%%%%%%%%%%%%%%%%%%%%%%%%%%%%

% \subsubsection{Uncertainty for Hallucination Detection}
\textbf{Uncertainty for Hallucination Detection. }
% \textbf{RQ1: Does SCoOP's aggregated uncertainty better detect hallucinations than aggregation baselines?}
We evaluate hallucination detection on 3-VLM systems comprising extra-large parameter models on ScienceQA benchmark. 
Performance is measured with the AUROC, as shown in the left panel of Fig.~\ref{fig:result_1}.
SCoOP attains the highest AUROC of 0.866, outperforming Majority Voting (0.757) by 10.9\% and Naive Selection (0.732) by 13.4\%. 
These gains confirm that SCoOP's aggregated uncertainty provides markedly superior separation of hallucinatory and correct responses compared to aggregation baselines.
For context, the green dashed line indicates the mean AUROC of the three individual VLMs without aggregation (0.789).
Both aggregation baselines, Majority Voting and Naive Selection, degrade relative to this single-model average by 3.2\% and 5.7\%, respectively. 
In contrast, SCoOP improves by 7.7\%. 
This demonstrates SCoOP's robustness to the elevated uncertainty that arises when aggregating different heterogeneous VLMs.
The superiority of SCoOP stems from its uncertainty-weighted pooling mechanism, which preserves informative disagreement among models rather than diluting it through unweighted voting or selection. 
Consequently, SCoOP yields more accurate uncertainty estimates for downstream hallucination detection.

% \subsubsection{Uncertainty for Abstention}
\textbf{Uncertainty for Abstention. }
% \textbf{RQ2: How well can SCoOP's estimated uncertainty lead to accuracy improvements when rejecting uncertain samples?}
We assess the practical utility of aggregated system-level uncertainty by measuring AURAC (right panel of Fig.~\ref{fig:result_1}), which quantifies accuracy gains achievable through abstention of high-uncertainty predictions.
SCoOP achieves the highest AURAC of 0.907, surpassing Majority Voting (0.840) by 6.7\% and Naive Selection (0.818) by 8.9\%. 
This demonstrates that SCoOP's uncertainty estimates are not only discriminative (high AUROC) but also well-estimated with respect to prediction reliability, which enables substantial accuracy improvements when high uncertain samples are rejected.
For reference, the green dashed line marks the mean AURAC of the three individual VLMs without aggregation (0.853), where uncertainty is computed similar to the preceding section. 
Both Majority Voting and Naive Selection fall below this single-model average by 1.3\% and 3.5\%, respectively, which reveals that unweighted or heuristic aggregation degrades the impacts of uncertainty. 
In contrast, SCoOP improves by 5.4\% over the single-model baseline, confirming that its uncertainty-weighted fusion enhances reliability signals beyond what any individual VLM provides.
These results collectively validate that SCoOP produces uncertainty estimates that are both discriminative for hallucination detection and useful for abstention, outperforming basic aggregation strategies across both metrics.

% \subsubsection{Aggregation Efficiency and Performance Trade-offs}
\textbf{Aggregation Efficiency and Performance Trade-offs. }
% \textbf{RQ5: What is the computational efficiency trade-off of SCoOP compared to baselines?} 
% Having established SCoOP's superior hallucination detection (AUROC) and abstention (AURAC) on ScienceQA (Fig.~\ref{fig:result_1}), we evaluate its aggregation efficiency using \emph{E2E-Latency@p50} with the same settings.
We further evaluate SCoOP's aggregation efficiency using \emph{E2E-Latency@p50} with the same settings as in Fig.~\ref{fig:result_1}.
Results are reported in Table~\ref{tab:efficiency}. 
Applying SCoOP in multi-VLM systems yields substantial improvements in hallucination detection (AUROC) and abstention (AURAC), while introducing less than one second of additional latency (0.05–0.85 $sec$) compared to systems without aggregation.
% For each system size, we list SCoOP's absolute latency $L_{\text{sec}}$ and the relative difference $\Delta L_{\mu s}$ (microseconds) of each baseline vs. SCoOP (negative values indicate the baseline is faster).
Compared to baseline aggregation methods, SCoOP incurs only microsecond-level overhead: it is slower than Majority Voting by 0.88–12.76 $\mu s$ and slower than Naive Selection by 2.98–12.14 $\mu s$.
Such differences are negligible in real-world deployments, particularly given that VLMs inference typically operates at the scale of seconds per sample.
Moreover, despite this minimal latency, SCoOP achieves significantly superior uncertainty quality and impact, as shown in Fig.\ref{fig:result_1}.
For example, the 3-VLM systems latency reported in Table~\ref{tab:efficiency}, SCoOP runs at 1.58 $sec$ per sample, which is only 1.87 $\mu s$ slower than Majority Voting and 7.88 $\mu s$ slower than Naive Selection. 
In light of its 10.9\% gain in AUROC and 6.7\% gain in AURAC over Majority Voting (Fig.~\ref{fig:result_1}), this microsecond-level overhead is easily justified in safety-critical or high-stakes applications.
Overall, SCoOP achieves superior uncertainty quality with negligible additional latency, making it highly practical for real-world multi-VLM systems.

%%%%%%%%%%%%%%%%%%%%%%%%%%%%%%%%%%%%%%%%%%%%%%%%%%%%%%%%%%%%%%%%%%%%%%%%%%%%%%%%%%%%%%%%%%%%%%%%
\begin{figure}[t]
    \centering
    \begin{minipage}[t]{0.50\columnwidth}
        \vspace{0pt}
        \centering
        \scriptsize
        \setlength{\tabcolsep}{4pt}
        \renewcommand{\arraystretch}{0.9}
        \captionof{table}{
        \footnotesize
        \textbf{Comparison of E2E-Latency@p50 across varying system sizes.}
        $L_{sec}$ denotes latency per sample (in seconds). 
        $\Delta L_{\mu s}$ denotes latency difference relative to SCoOP (in microseconds).
        Negative values indicate faster than SCoOP.
        }
        \vspace{-5pt}
        \resizebox{\linewidth}{!}{
        \begin{tabular}{l|c|ccc}
        \toprule
        \textbf{\#VLMs} &
        \makecell{\textbf{No}\\ \textbf{Aggregation}\\[-1pt]\scriptsize{($L_{sec}\downarrow$)}} &
        \makecell{\textbf{SCoOP}\\ \textbf{(ours)}\\[-1pt]\scriptsize{($L_{sec}\downarrow$)}} &
        \makecell{\textbf{Majority}\\ \textbf{Voting}\\[-1pt]\scriptsize{($\Delta L_{\mu s}\downarrow$)}} &
        \makecell{\textbf{Naive}\\ \textbf{Selection}\\[-1pt]\scriptsize{($\Delta L_{\mu s}\downarrow$)}} \\
        \midrule
        2-VLM & $0.97$ & $1.02$ & $-0.88$ & $-9.86$ \\
        3-VLM & $1.45$ & $1.58$ & $-1.87$ & $-7.88$ \\
        4-VLM & $1.94$ & $2.38$ & $-4.27$ & $-12.14$ \\
        5-VLM & $2.42$ & $3.27$ & $-12.76$ & $-2.98$ \\
        \bottomrule
        \end{tabular}
        }
        \label{tab:efficiency}
    \end{minipage}
    \hfill
    \begin{minipage}[t]{0.46\columnwidth}
        \vspace{0pt}
        \centering
        \scriptsize
        \setlength{\tabcolsep}{3pt}
        \captionof{table}{
        \footnotesize
        \textbf{Accuracy (\%) across model parameter scales on ScienceQA.} 
        \emph{Avg. of Single Models} is for reference (mean accuracy of individual models), not an aggregation method. 
        Within the three aggregation methods, \textbf{bold} marks the highest score per row and \underline{underline} marks the second highest.
        }
        \vspace{-5pt}
        % \vspace{4pt}
        \resizebox{\linewidth}{!}{%
        \begin{tabular}{lccc|c}
        \toprule
        \makecell{\textbf{Param.}\\ \textbf{Scale}} & 
        \makecell{\textbf{SCoOP}\\ \textbf{(ours)}} & 
        \makecell{\textbf{Majority}\\ \textbf{Voting}} & 
        \makecell{\textbf{Naive}\\ \textbf{Selection}} & 
        \makecell{\textbf{Avg. of}\\\textbf{Single Models}} \\
        \midrule
        Small        & \underline{$67.33$} & $\mathbf{67.44}$ & $66.49$ & $62.95$ \\
        Medium       & \underline{$71.32$} & $\mathbf{71.53}$ & $69.43$ & $66.05$ \\
        Large        & \underline{$70.06$} & $\mathbf{70.69}$ & $68.91$ & $69.16$ \\
        Extra Large  & \underline{$71.85$} & $71.11$ & $\mathbf{72.90}$ & $69.75$ \\
        \bottomrule
        \end{tabular}%
        }
        \label{tab:acc_by_scale}
    \end{minipage}
    \vspace{-10pt}
\end{figure}
%%%%%%%%%%%%%%%%%%%%%%%%%%%%%%%%%%%%%%%%%%%%%%%%%%%%%%%%%%%%%%%%%%%%%%%%%%%%%%%%%%%%%%%%%%%%%%%%
\vspace{-3pt}

\subsection{Research Questions and Ablation Study}

\vspace{-3pt}

%%%%%%%%%%%%%%%%%%%%%%%%%%%%%%%%%%%%%%%%%%%%%%%%%%%%%%%%%%%%%%%%%%%%%%%%%%%%%%%%%%%%%%%%%%%%%%%%
% \begin{figure}[t]
%     \centering
%     \vspace{-10pt}
%     \begin{subfigure}[t]{0.48\columnwidth}
%         \centering
%         \includegraphics[width=\linewidth]{img/result_4.png}
%         \caption{AUROC for hallucination detection.}
%         \label{fig:result_4}
%     \end{subfigure}
%     \vspace{-10pt}
%     \hfill
%     \begin{subfigure}[t]{0.48\columnwidth}
%         \centering
%         \includegraphics[width=\linewidth]{img/result_5.png}
%         \caption{AURAC for abstention.}
%         \label{fig:result_5}
%     \end{subfigure}
%     \caption{
%         \footnotesize{
%         \textbf{UQ performance across varying model parameter scales.}
%         SCoOP (ours), Majority Voting, and Naive Selection are evaluated in 3-VLM systems on ScienceQA.
%         % (a) reflects hallucination detection quality, while (b) shows AURAC captures accuracy improvements by rejecting uncertain samples.
%         }
%     }
%     \label{fig:model_size_scaling}
%     \vspace{-15pt}
% \end{figure}
%%%%%%%%%%%%%%%%%%%%%%%%%%%%%%%%%%%%%%%%%%%%%%%%%%%%%%%%%%%%%%%%%%%%%%%%%%%%%%%%%%%%%%%%%%%%%%%%

\begin{wrapfigure}{r}{0.45\columnwidth}
    \vspace{-20pt}
    \centering
    \includegraphics[width=1\linewidth]{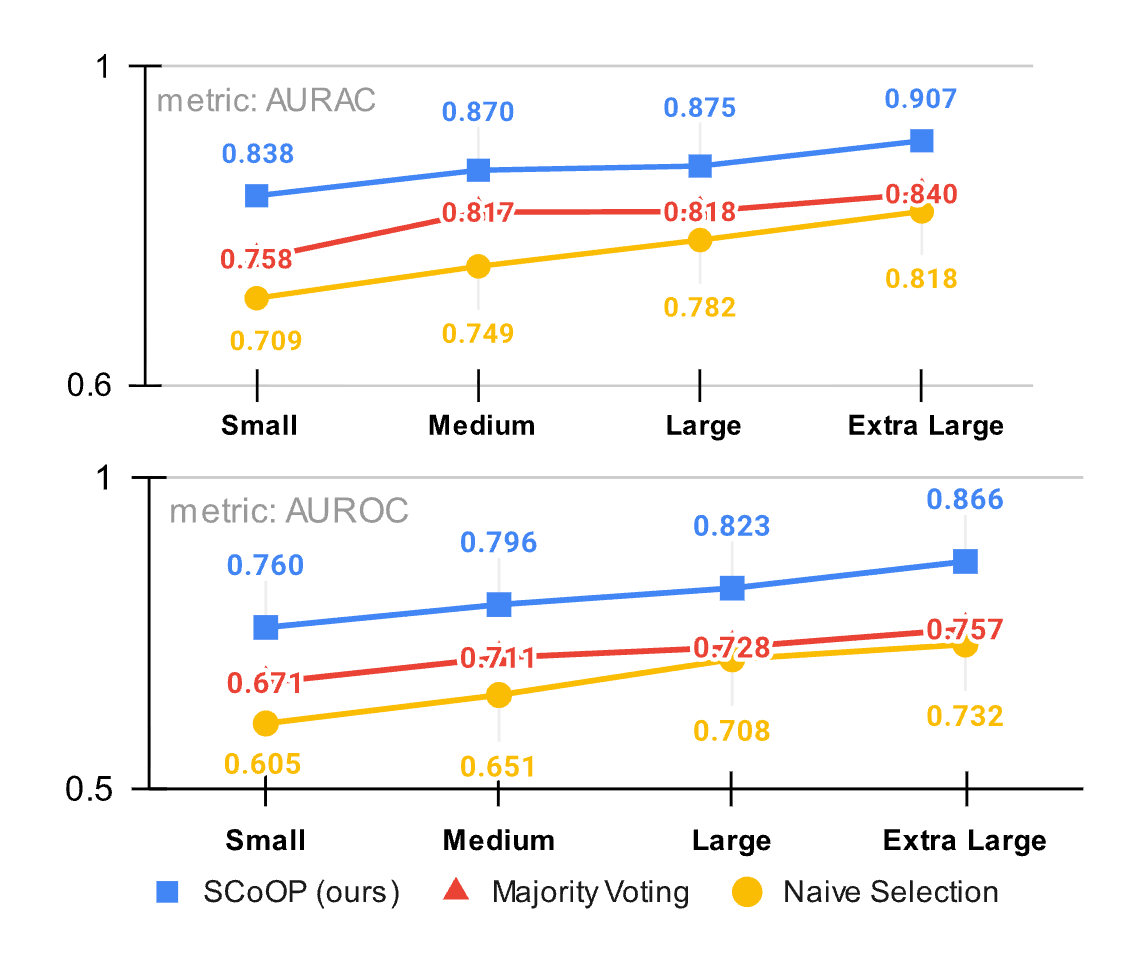}
    \vspace{-20pt}
    \caption{
        \footnotesize{
        \textbf{UQ performance across varying model parameter scales.}
        Methods are evaluated in 3-VLM systems on ScienceQA.
        AUROC for hallucination detection; AURAC for abstention.
        % (a) reflects hallucination detection quality, while (b) shows AURAC captures accuracy improvements by rejecting uncertain samples.
        }
    }
    \label{fig:model_size_scaling}
    \vspace{-13pt}
\end{wrapfigure}

\textbf{RQ1: How does the model parameter scale impact UQ performance?}
We evaluate 3-VLM systems across four parameter scales (small, medium, large, extra large), covering 19 unique combinations, on ScienceQA as shown in Fig.~\ref{fig:model_size_scaling}.
SCoOP consistently achieves the highest performance across all scales.
It outperforms Majority Voting by 8.9\% to 10.9\% and Naive Selection by 13.4\% to 15.5\%.  
It surpasses Majority Voting by 8.0\% to 8.9\% and Naive Selection by 9.0\% to 12.9\%.
All methods improve with scale, but SCoOP maintains dominant gains at every level. 
These results demonstrate that SCoOP generalizes robustly across model sizes, scaling effectively from small to extra large VLMs while preserving superior hallucination detection and abstention.

%%%%%%%%%%%%%%%%%%%%%%%%%%%%%%%%%%%%%%%%%%%%%%%%%%%%%%%%%%%%%%%%%%%%%%%%%%%%%%%%%%%%%%%%%%%%%%%%
% \begin{figure}[t]
%     \centering

%     % Subfigure (a)
%     \begin{subfigure}[t]{0.48\columnwidth}
%         \centering
%         \includegraphics[width=\linewidth]{img/result_2.png}
%         % \caption{AUROC for hallucination detection.}
%         \label{fig:result_2}
%     \end{subfigure}
%     \hfill
%     % Subfigure (b)
%     \begin{subfigure}[t]{0.48\columnwidth}
%         \centering
%         \includegraphics[width=\linewidth]{img/result_3.png}
%         % \caption{AURAC for abstention.}
%         \label{fig:result_3}
%     \end{subfigure}

%     \caption{
%         \footnotesize
%         \textbf{UQ performance across varying system sizes on ScienceQA.}
%         Methods are evaluated across various 2- to 5-VLM systems using large-scale models. AUROC for hallucination detection; AURAC for abstention.
%         % Lighter to darker bars represent increasing numbers of models in the system, and the cyan dashed line indicates the average performance of each method.
%     }
%     \label{fig:result_scaling}
%     \vspace{-10pt}
% \end{figure}
%%%%%%%%%%%%%%%%%%%%%%%%%%%%%%%%%%%%%%%%%%%%%%%%%%%%%%%%%%%%%%%%%%%%%%%%%%%%%%%%%%%%%%%%%%%%%%%%

%%%%%%%%%%%%%%%%%%%%%%%%%%%%%%%%%%%%%%%%%%%%%%%%%%%%%%%%%%%%%%%%%%%%%%%%%%%%%%%%%%%%%%%%%%%%%%%%
%%%%%%%%%%%%%%%%%%%%%%%%%%%%%%%%%%%%%%%%%%%%%%%%%%%%%%%%%%%%%%%%%%%%%%%%%%%%%%%%%%%%%%%%%%%%%%%%

\begin{figure}[t]
    \centering
    \includegraphics[width=1\linewidth]{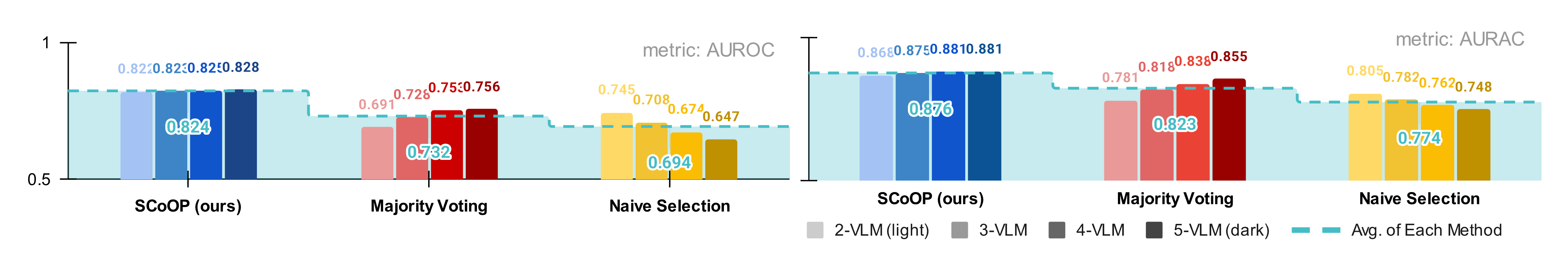}
    \vspace{-25pt}
    \caption{
        \footnotesize{
        \textbf{UQ performance across varying system sizes on ScienceQA.}
        Methods are evaluated across various 2- to 5-VLM systems using large-scale models. 
        AUROC for hallucination detection; AURAC for abstention.
        }
    }
    \label{fig:result_scaling}
    \vspace{-10pt}
\end{figure}

\textbf{RQ2: How does the system size impact UQ performance?}
We evaluate how system size $k \in \{2,3,4,5\}$ affects uncertainty estimates, yielding 26 unique $k$-VLM configurations. 
All results are reported in Fig.~\ref{fig:result_scaling}.
SCoOP achieves the highest performance across all configurations.
It attains a mean of 0.824 AUROC across different system sizes, which outperforms Majority Voting (0.732) by 9.2\% and Naive Selection (0.694) by 13\%.
It reaches a mean of 0.876 AURAC surpassing Majority Voting (0.823) by 5.3\% and Naive Selection (0.774) by 10.2\%.
Moreover, SCoOP maintains near-perfect stability: AUROC varies by $\leq 0.6\%$, and AURAC by $\leq 1.3\%$ across $k$ VLMs. 
In contrast, Majority Voting improves with $k$, while Naive Selection degrades.
These results confirm that SCoOP's uncertainty-weighted pooling robustly preserves high-quality uncertainty signals regardless of system size, outperforming unweighted and heuristic baselines across all 26 configurations.

%%%%%%%%%%%%%%%%%%%%%%%%%%%%%%%%%%%%%%%%%%%%%%%%%%%%%%%%%%%%%%%%%%%%%%%%%%%%%%%%%%%%%%%%%%%%%%%%

%%%%%%%%%%%%%%%%%%%%%%%%%%%%%%%%%%%%%%%%%%%%%%%%%%%%%%%%%%%%%%%%%%%%%%%%%%%%%%%%%%%%%%%%%%%%%%%%

\textbf{RQ3: Can SCoOP improve the benchmark accuracy via aggregation?}
While SCoOP is designed primarily for quantifying uncertainty, it does not sacrifice accuracy and often improves it relative to using any single model. 
As shown in Table~\ref{tab:acc_by_scale}, SCoOP consistently outperforms the \emph{Avg. of Single Models} reference (gains range from 0.9\% to 5.3\%), indicating that its uncertainty-weighted pooling extracts complementary signal from heterogeneous VLMs. 
Compared to aggregation baselines, SCoOP is competitive with Majority Voting and remains close to the best observed accuracy even when Naive Selection attains a slight edge on the extra large scale.
Taken together, these results show that SCoOP's uncertainty-aware fusion can deliver accuracy at least on par with strong baselines while simultaneously yielding much superior uncertainty estimates, making it a favorable aggregation choice when both reliability and accuracy matter.

% \subsection{Robustness Under Challenging Conditions}
\textbf{RQ4: Robustness Under Challenging Conditions. }
% \textbf{RQ6: Can SCoOP provide informative uncertainty under low-accuracy conditions?}
We evaluate uncertainty quality under challenging conditions where all individual VLMs obtain $<$ 50\% accuracy on the benchmark. 
Such regimes stress-test aggregation methods, as raw predictions are unreliable and uncertainty signals are prone to degradation.
Results are reported in Table~\ref{tab:combined_lowacc}.
On MMBench, SCoOP reaches 0.608 (4-VLM), outperforming Majority Voting (0.532) by 7.6\% and Naive Selection (0.533) by 7.5\%.  
On MMMU, SCoOP attains 0.667 (4-VLM), surpassing Majority Voting (0.583) by 8.4\% and Naive Selection (0.576) by 9.1\%.
Both baselines degrade to near non-informative (AUROC close to 0.5), while SCoOP maintains informative uncertainty where unweighted or heuristic methods fail.
These results confirm that SCoOP extracts reliable confidence signals even from highly unreliable models, enabling robust hallucination detection in low-accuracy for real-world deployment scenarios.

%%%%%%%%%%%%%%%%%%%%%%%%%%%%%%%%%%%%%%%%%%%%%%%%%%%%%%%%%%%%%%%%%%%%%%%%%%%%%%%%%%%%%%%%%%%%%%%%
%%%%%%%%%%%%%%%%%%%%%%%%%%%%%%%%%%%%%%%%%%%%%%%%%%%%%%%%%%%%%%%%%%%%%%%%%%%%%%%%%%%%%%%%%%%%%%%%

\begin{table}[h]
\centering
\scriptsize
\setlength{\tabcolsep}{4pt}
\caption{\footnotesize{\textbf{AUROC under conditions where all individual VLMs yield low benchmark accuracy ($<$50\%).} 
Each configuration corresponds to $K$-VLM systems with medium scale models.
\textbf{Bold} values indicate the highest score for each VLM system.
\underline{Underline} values indicate the second highest score.}}
\vspace{5pt}

\begin{subtable}{0.48\linewidth}
    \centering
    \begin{tabular}{lccc}
        \toprule
        \multicolumn{4}{c}{\textbf{MMBench}} \\
        \midrule
        \textbf{\#VLMs} & \textbf{SCoOP (ours)} & \textbf{Majority Voting} & \textbf{Naive Selection} \\
        \midrule
        2-VLM & $\mathbf{0.586}$ & $0.553$ & \underline{$0.555$} \\
        3-VLM & $\mathbf{0.587}$ & \underline{$0.549$} & $0.545$ \\
        4-VLM & $\mathbf{0.608}$ & $0.532$ & \underline{$0.533$} \\
        \bottomrule
    \end{tabular}
    \label{tab:lowacc_mmbench}
\end{subtable}
\hfill
\begin{subtable}{0.48\linewidth}
    \centering
    \begin{tabular}{lccc}
        \toprule
        \multicolumn{4}{c}{\textbf{MMMU}} \\
        \midrule
        \textbf{\#VLMs} & \textbf{SCoOP (ours)} & \textbf{Majority Voting} & \textbf{Naive Selection} \\
        \midrule
        2-VLM & $\mathbf{0.627}$ & $0.530$ & \underline{$0.615$} \\
        3-VLM & $\mathbf{0.646}$ & $0.554$ & \underline{$0.602$} \\
        4-VLM & $\mathbf{0.667}$ & \underline{$0.583$} & $0.576$ \\
        \bottomrule
    \end{tabular}
    \label{tab:lowacc_mmmu}
\end{subtable}
% \vspace{-10pt}

\label{tab:combined_lowacc}
\end{table}
\section{Conclusion}
\label{sec:conclusion}
\vspace{-5pt}

We presented \textbf{SCoOP}, a training-free uncertainty quantification (UQ) framework that aggregates the probabilistic opinions of multiple VLMs through uncertainty-weighted linear opinion pooling to estimate system-level uncertainty in multi-VLM systems. 
It demonstrates strong hallucination detection and abstention performance (AUROC $=0.866$, AURAC $=0.907$), consistently outperforming Majority Voting and Naive Selection baselines. 
Importantly, SCoOP achieves these gains with only microsecond-level latency overhead relative to the baselines, which is insignificant compared to the typical VLMs inference time (on the order of seconds). 
This shows that improved uncertainty quality can be achieved without compromising system efficiency.
Overall, SCoOP establishes an effective and efficient approach to quantify uncertainty in multi-VLM systems -- providing reliable hallucination detection and abstention with minimal aggregation overhead, advancing the reliability and deployability of multimodal AI systems.

\textbf{Limitations.} While the aggregation itself is computationally inexpensive, the total inference latency is still dominated by per-model sampling, which scales with the number of models ($K$) and the number of samples per model ($N$). 
Although this remains a limiting factor, SCoOP's computational lightweight aggregation provides a practical foundation for scalable uncertainty-aware multi-VLM systems.
\textbf{Future directions.} We aim to expand SCoOP beyond multiple-choice to free-form Visual Question Answering (VQA) and integrating it with agentic routing or multi-step reasoning pipelines to leverage uncertainty for adaptive decision-making in multi-agent systems. 

% \newpage

\section*{Acknowledgments}
This work was supported in part by the U.S. Military Academy (USMA) under Cooperative Agreement No. W911NF-23-2-0108. The views and conclusions expressed in this paper are those of the authors and do not reflect the official policy or position of the U.S. Military Academy, U.S. Army, U.S. Department of Defense, or U.S. Government.

%Bibliography
\bibliographystyle{unsrt}  
\bibliography{references}  

@article{lu2022learn,
  title={Learn to explain: Multimodal reasoning via thought chains for science question answering},
  author={Lu, Pan and Mishra, Swaroop and Xia, Tanglin and Qiu, Liang and Chang, Kai-Wei and Zhu, Song-Chun and Tafjord, Oyvind and Clark, Peter and Kalyan, Ashwin},
  journal={Advances in Neural Information Processing Systems},
  volume={35},
  pages={2507--2521},
  year={2022}
}

@inproceedings{yue2024mmmu,
  title={Mmmu: A massive multi-discipline multimodal understanding and reasoning benchmark for expert agi},
  author={Yue, Xiang and Ni, Yuansheng and Zhang, Kai and Zheng, Tianyu and Liu, Ruoqi and Zhang, Ge and Stevens, Samuel and Jiang, Dongfu and Ren, Weiming and Sun, Yuxuan and others},
  booktitle={Proceedings of the IEEE/CVF Conference on Computer Vision and Pattern Recognition},
  pages={9556--9567},
  year={2024}
}

@inproceedings{liu2024mmbench,
  title={Mmbench: Is your multi-modal model an all-around player?},
  author={Liu, Yuan and Duan, Haodong and Zhang, Yuanhan and Li, Bo and Zhang, Songyang and Zhao, Wangbo and Yuan, Yike and Wang, Jiaqi and He, Conghui and Liu, Ziwei and others},
  booktitle={European conference on computer vision},
  pages={216--233},
  year={2024},
  organization={Springer}
}

@article{liu2024survey,
  title={A survey on hallucination in large vision-language models},
  author={Liu, Hanchao and Xue, Wenyuan and Chen, Yifei and Chen, Dapeng and Zhao, Xiutian and Wang, Ke and Hou, Liping and Li, Rongjun and Peng, Wei},
  journal={arXiv preprint arXiv:2402.00253},
  year={2024}
}

@inproceedings{li2025survey,
  title={A Survey of State of the Art Large Vision Language Models: Benchmark Evaluations and Challenges},
  author={Li, Zongxia and Wu, Xiyang and Du, Hongyang and Liu, Fuxiao and Nghiem, Huy and Shi, Guangyao},
  booktitle={Proceedings of the Computer Vision and Pattern Recognition Conference},
  pages={1587--1606},
  year={2025}
}

@inproceedings{liu2025uncertainty,
  title={Uncertainty quantification and confidence calibration in large language models: A survey},
  author={Liu, Xiaoou and Chen, Tiejin and Da, Longchao and Chen, Chacha and Lin, Zhen and Wei, Hua},
  booktitle={Proceedings of the 31st ACM SIGKDD Conference on Knowledge Discovery and Data Mining V. 2},
  pages={6107--6117},
  year={2025}
}

@article{shorinwa2025survey,
  title={A survey on uncertainty quantification of large language models: Taxonomy, open research challenges, and future directions},
  author={Shorinwa, Ola and Mei, Zhiting and Lidard, Justin and Ren, Allen Z and Majumdar, Anirudha},
  journal={ACM Computing Surveys},
  year={2025},
  publisher={ACM New York, NY}
}

@article{fadeeva2024fact,
  title={Fact-checking the output of large language models via token-level uncertainty quantification},
  author={Fadeeva, Ekaterina and Rubashevskii, Aleksandr and Shelmanov, Artem and Petrakov, Sergey and Li, Haonan and Mubarak, Hamdy and Tsymbalov, Evgenii and Kuzmin, Gleb and Panchenko, Alexander and Baldwin, Timothy and others},
  journal={arXiv preprint arXiv:2403.04696},
  year={2024}
}

@article{ling2024uncertainty,
  title={Uncertainty decomposition and quantification for in-context learning of large language models},
  author={Ling, Chen and Zhao, Xujiang and Cheng, Wei and Liu, Yanchi and Sun, Yiyou and Zhang, Xuchao and Oishi, Mika and Osaki, Takao and Matsuda, Katsushi and Ji, Jie and others},
  journal={CoRR},
  year={2024}
}

@article{wang2025predictive,
  title={Predictive Auditing of Hidden Tokens in LLM APIs via Reasoning Length Estimation},
  author={Wang, Ziyao and Sun, Guoheng and He, Yexiao and Shen, Zheyu and Tian, Bowei and Li, Ang},
  journal={arXiv preprint arXiv:2508.00912},
  year={2025}
}

@article{finlayson2024logits,
  title={Logits of api-protected llms leak proprietary information},
  author={Finlayson, Matthew and Ren, Xiang and Swayamdipta, Swabha},
  journal={arXiv preprint arXiv:2403.09539},
  year={2024}
}

@article{lin2022teaching,
  title={Teaching models to express their uncertainty in words},
  author={Lin, Stephanie and Hilton, Jacob and Evans, Owain},
  journal={arXiv preprint arXiv:2205.14334},
  year={2022}
}

@article{stengel2024lacie,
  title={LACIE: Listener-aware finetuning for calibration in large language models},
  author={Stengel-Eskin, Elias and Hase, Peter and Bansal, Mohit},
  journal={Advances in Neural Information Processing Systems},
  volume={37},
  pages={43080--43106},
  year={2024}
}

@article{lin2023generating,
  title={Generating with confidence: Uncertainty quantification for black-box large language models},
  author={Lin, Zhen and Trivedi, Shubhendu and Sun, Jimeng},
  journal={arXiv preprint arXiv:2305.19187},
  year={2023}
}

@article{farquhar2024detecting,
  title={Detecting hallucinations in large language models using semantic entropy},
  author={Farquhar, Sebastian and Kossen, Jannik and Kuhn, Lorenz and Gal, Yarin},
  journal={Nature},
  volume={630},
  number={8017},
  pages={625--630},
  year={2024},
  publisher={Nature Publishing Group UK London}
}

@article{xiong2023can,
  title={Can llms express their uncertainty? an empirical evaluation of confidence elicitation in llms},
  author={Xiong, Miao and Hu, Zhiyuan and Lu, Xinyang and Li, Yifei and Fu, Jie and He, Junxian and Hooi, Bryan},
  journal={arXiv preprint arXiv:2306.13063},
  year={2023}
}

@article{su2024api,
  title={Api is enough: Conformal prediction for large language models without logit-access},
  author={Su, Jiayuan and Luo, Jing and Wang, Hongwei and Cheng, Lu},
  journal={arXiv preprint arXiv:2403.01216},
  year={2024}
}

@inproceedings{ji2023map,
  title={Map: Multimodal uncertainty-aware vision-language pre-training model},
  author={Ji, Yatai and Wang, Junjie and Gong, Yuan and Zhang, Lin and Zhu, Yanru and Wang, Hongfa and Zhang, Jiaxing and Sakai, Tetsuya and Yang, Yujiu},
  booktitle={Proceedings of the IEEE/CVF conference on computer vision and pattern recognition},
  pages={23262--23271},
  year={2023}
}

@article{chen2024unveiling,
  title={Unveiling uncertainty: A deep dive into calibration and performance of multimodal large language models},
  author={Chen, Zijun and Hu, Wenbo and He, Guande and Deng, Zhijie and Zhang, Zheng and Hong, Richang},
  journal={arXiv preprint arXiv:2412.14660},
  year={2024}
}

@article{zhang2024vl,
  title={Vl-uncertainty: Detecting hallucination in large vision-language model via uncertainty estimation},
  author={Zhang, Ruiyang and Zhang, Hu and Zheng, Zhedong},
  journal={arXiv preprint arXiv:2411.11919},
  year={2024}
}

@article{fang2024uncertainty,
  title={From uncertainty to trust: Enhancing reliability in vision-language models with uncertainty-guided dropout decoding},
  author={Fang, Yixiong and Yang, Ziran and Chen, Zhaorun and Zhao, Zhuokai and Zhou, Jiawei},
  journal={arXiv preprint arXiv:2412.06474},
  year={2024}
}

@article{li2025mitigating,
  title={Mitigating Hallucinations in Large Vision-Language Models via Reasoning Uncertainty-guided Refinement},
  author={Li, Shenshen and Xu, Xing and Meng, Wenxin and Song, Jingkuan and Peng, Chong and Shen, Heng Tao},
  journal={IEEE Transactions on Multimedia},
  year={2025},
  publisher={IEEE}
}

@article{zhi2025seeing,
  title={Seeing and Reasoning with Confidence: Supercharging Multimodal LLMs with an Uncertainty-Aware Agentic Framework},
  author={Zhi, Zhuo and Feng, Chen and Daneshmend, Adam and Orlu, Mine and Demosthenous, Andreas and Yin, Lu and Li, Da and Liu, Ziquan and Rodrigues, Miguel RD},
  journal={arXiv preprint arXiv:2503.08308},
  year={2025}
}

@article{han2024towards,
  title={Towards uncertainty-aware language agent},
  author={Han, Jiuzhou and Buntine, Wray and Shareghi, Ehsan},
  journal={arXiv preprint arXiv:2401.14016},
  year={2024}
}

@article{liu2025advances,
  title={Advances and challenges in foundation agents: From brain-inspired intelligence to evolutionary, collaborative, and safe systems},
  author={Liu, Bang and Li, Xinfeng and Zhang, Jiayi and Wang, Jinlin and He, Tanjin and Hong, Sirui and Liu, Hongzhang and Zhang, Shaokun and Song, Kaitao and Zhu, Kunlun and others},
  journal={arXiv preprint arXiv:2504.01990},
  year={2025}
}

@article{xie2024large,
  title={Large multimodal agents: A survey},
  author={Xie, Junlin and Chen, Zhihong and Zhang, Ruifei and Wan, Xiang and Li, Guanbin},
  journal={arXiv preprint arXiv:2402.15116},
  year={2024}
}

@article{lu2024merge,
  title={Merge, ensemble, and cooperate! a survey on collaborative strategies in the era of large language models},
  author={Lu, Jinliang and Pang, Ziliang and Xiao, Min and Zhu, Yaochen and Xia, Rui and Zhang, Jiajun},
  journal={arXiv preprint arXiv:2407.06089},
  year={2024}
}

@inproceedings{pan2025multiagent,
  title={Why do multiagent systems fail?},
  author={Pan, Melissa Z and Cemri, Mert and Agrawal, Lakshya A and Yang, Shuyi and Chopra, Bhavya and Tiwari, Rishabh and Keutzer, Kurt and Parameswaran, Aditya and Ramchandran, Kannan and Klein, Dan and others},
  booktitle={ICLR 2025 Workshop on Building Trust in Language Models and Applications},
  year={2025}
}

@inproceedings{duan2025enhancing,
  title={Enhancing multi-agent consensus through third-party llm integration: Analyzing uncertainty and mitigating hallucinations in large language models},
  author={Duan, Zhihua and Wang, Jialin},
  booktitle={2025 8th International Conference on Advanced Algorithms and Control Engineering (ICAACE)},
  pages={2222--2227},
  year={2025},
  organization={IEEE}
}

@article{chen2025harnessing,
  title={Harnessing multiple large language models: A survey on llm ensemble},
  author={Chen, Zhijun and Li, Jingzheng and Chen, Pengpeng and Li, Zhuoran and Sun, Kai and Luo, Yuankai and Mao, Qianren and Yang, Dingqi and Sun, Hailong and Yu, Philip S},
  journal={arXiv preprint arXiv:2502.18036},
  year={2025}
}

@article{li2024more,
  title={More agents is all you need},
  author={Li, Junyou and Zhang, Qin and Yu, Yangbin and Fu, Qiang and Ye, Deheng},
  journal={arXiv preprint arXiv:2402.05120},
  year={2024}
}

@article{ai2025beyond,
  title={Beyond Majority Voting: LLM Aggregation by Leveraging Higher-Order Information},
  author={Ai, Rui and Pan, Yuqi and Simchi-Levi, David and Tambe, Milind and Xu, Haifeng},
  journal={arXiv preprint arXiv:2510.01499},
  year={2025}
}

@article{li2024reference,
  title={Reference-free hallucination detection for large vision-language models},
  author={Li, Qing and Geng, Jiahui and Lyu, Chenyang and Zhu, Derui and Panov, Maxim and Karray, Fakhri},
  journal={arXiv preprint arXiv:2408.05767},
  year={2024}
}

@inproceedings{xiao2025detecting,
  title={Detecting and mitigating hallucination in large vision language models via fine-grained ai feedback},
  author={Xiao, Wenyi and Huang, Ziwei and Gan, Leilei and He, Wanggui and Li, Haoyuan and Yu, Zhelun and Shu, Fangxun and Jiang, Hao and Zhu, Linchao},
  booktitle={Proceedings of the AAAI Conference on Artificial Intelligence},
  volume={39},
  pages={25543--25551},
  year={2025}
}

@article{xue2025verify,
  title={Verify when uncertain: Beyond self-consistency in black box hallucination detection},
  author={Xue, Yihao and Greenewald, Kristjan and Mroueh, Youssef and Mirzasoleiman, Baharan},
  journal={arXiv preprint arXiv:2502.15845},
  year={2025}
}

@article{padhi2025calibrating,
  title={Calibrating Uncertainty Quantification of Multi-Modal LLMs using Grounding},
  author={Padhi, Trilok and Kaur, Ramneet and Cobb, Adam D and Acharya, Manoj and Roy, Anirban and Samplawski, Colin and Matejek, Brian and Berenbeim, Alexander M and Bastian, Nathaniel D and Jha, Susmit},
  journal={arXiv preprint arXiv:2505.03788},
  year={2025}
}

@inproceedings{qu2025uq,
  title={UQ-Merge: Uncertainty Guided Multimodal Large Language Model Merging},
  author={Qu, Huaizhi and Zhao, Xinyu and Peng, Jie and Lee, Kwonjoon and Dariush, Behzad and Chen, Tianlong},
  booktitle={Findings of the Association for Computational Linguistics: ACL 2025},
  pages={1401--1417},
  year={2025}
}

@article{zhang2025consensus,
  title={Consensus Entropy: Harnessing Multi-VLM Agreement for Self-Verifying and Self-Improving OCR},
  author={Zhang, Yulong and Liang, Tianyi and Huang, Xinyue and Cui, Erfei and Guo, Xu and Chu, Pei and Li, Chenhui and Zhang, Ru and Wang, Wenhai and Liu, Gongshen},
  journal={arXiv preprint arXiv:2504.11101},
  year={2025}
}

@article{carvalho2012consensual,
  title={A consensual linear opinion pool},
  author={Carvalho, Arthur and Larson, Kate},
  journal={arXiv preprint arXiv:1204.5399},
  year={2012}
}

@article{stone1961opinion,
  title={The opinion pool},
  author={Stone, Mervyn},
  journal={The Annals of Mathematical Statistics},
  pages={1339--1342},
  year={1961},
  publisher={JSTOR}
}

\clearpage
\appendix
\onecolumn

\section{Appendix}

%%%%%%%%%%%%%%%%%%%%%%%%%%%%%%%%%%%%%%%%%%%%%%%%%%%%%%%%%%%%
%%%%%%%%%%%%%%%%%%%%%%%%%%%%%%%%%%%%%%%%%%%%%%%%%%%%%%%%%%%%

\subsection{Additional Math Details of SCoOP}
\label{sec:appendix_a}

\textbf{Notation Table. }
Table~\ref{tab:notation_scope} provides a summary of the mathematical symbols and notations used throughout this paper.

\begin{table*}[h]
\centering
\footnotesize
\caption{Notation summary for the SCoOP framework.}
\vspace{-5pt}
\begin{tabular}{ll}
\toprule
\textbf{Symbol} & \textbf{Description} \\
\midrule
$K$ & Number of VLMs (models or experts) in the system \\
$M_k$ & The $k$-th VLM \\
$\mathcal{M}=\{M_1,\ldots,M_K\}$ & Multi-VLM system composed of $K$ models \\
$x=(I,q)$ & Multiple-choice question (MCQ) input with image $I$ and text question $q$ \\
$N$ & Number of sampled responses per model \\
$y_k^n$ & $n$-th raw response sampled from model $M_k$ \\
$Y_k=\{y_k^1,\ldots,y_k^N\}$ & Set of $N$ sampled responses from model $M_k$ \\
$\Theta=\{\theta^{1}, \ldots, \theta^{J}\}$ & Set of all discrete answer options for the MCQ (size $J$) \\
$p_k(\theta^j \mid x)$ & Empirical probability that $M_k$ selects option $\theta^j$ for input $x$ \\
$\mathbf{p}_k$ & Probability vector $\big[p_k(\theta^1),\ldots,p_k(\theta^{J})\big]$ of model $M_k$ over $\Theta$ \\
$H_{k}$ & Shannon entropy of model $M_k$'s probability distribution $\mathbf{p}_k$ \\
$\mathrm{conf}_{k}$ & Confidence score of model $M_k$, defined as $1/H_{k}$ \\
$w_k$ & Normalized weight of model $M_k$ for opinion pooling \\
$\mathbf{p}_{\mathrm{agg}}$ & Aggregated probability vector from all models, $\sum_{k=1}^{K} w_{k} \mathbf{p}_{k}$ \\
$\theta^{*}$ & Final system prediction: the option with the highest aggregated probability \\
$H_{\mathrm{agg}}$ & Shannon entropy of the aggregated probability distribution $\mathbf{p}_{\mathrm{agg}}$ (System Uncertainty) \\
$H_{\mathrm{agg}}^{\mathrm{norm}}$ & Normalized system entropy, $H_{\mathrm{agg}} / \log_2 J$ \\
\bottomrule
\end{tabular}
\label{tab:notation_scope}
\end{table*}

\textbf{Pseudocode of SCoOP Algorithm. }
\label{app:algo}

%%%%%%%%%%%%%%%%%%%%%%%%%%%%%%%%%%%%%%%%%%%%%%%%%%%%%%%%%%%%%%%%%%%%%%
\begin{algorithm}[h]
\footnotesize
\caption{SCoOP: Semantic-Consistent Opinion Pooling for Uncertainty Quantification}
\label{alg:scoop}

\KwIn{$\mathcal{M}=\{M_1,\ldots,M_K\}$, input $x=(I,q)$, $\Theta=\{\theta^1,\ldots,\theta^J\}$, $N$ samples}
\KwOut{$\theta^*$ and $H_{\mathrm{agg}}^{\mathrm{norm}}$}

\BlankLine
\tcp{Step 1: Formulating single-VLM opinions}

\For{$k \leftarrow 1$ \KwTo $K$}{
    $Y_k \leftarrow$ sample $N$ responses from $M_k$ at temperature $T$\;

    % \tcp{Option A: use a matching function to avoid an overlong indicator line}
    % $p_k(\theta^j) \leftarrow \frac{1}{N}\sum_{n=1}^{N} \mathbb{I}\!\left[\textsc{Match}(y_k^n,\theta^j)\right]$\;
    $z_k^n \leftarrow \textsc{Match}(y_k^n, \Theta)$\;
    $p_k(\theta^j) \leftarrow \frac{1}{N}\sum_{n=1}^{N} \mathbb{I}[z_k^n = j]$\;
    $\mathbf{p}_k \leftarrow [p_k(\theta^1),\ldots,p_k(\theta^J)]$\;

    $H_k \leftarrow -\sum_{j=1}^{J} p_k(\theta^j)\log p_k(\theta^j)$\;

    $\mathrm{conf}_k \leftarrow 1/(H_k + \epsilon)$\;

}
$w_k \leftarrow \mathrm{conf}_k \big/ \sum_{k'=1}^{K} \mathrm{conf}_{k'}$\;

\BlankLine
\tcp{Step 2: Multi-VLM opinion aggregation}

$\mathbf{p}_{\mathrm{agg}} \leftarrow \sum_{k=1}^{K} w_k\,\mathbf{p}_k$\;

$\theta^* \leftarrow \arg\max_{\theta^j} p_{\mathrm{agg}}(\theta^j)$\;

$H_{\mathrm{agg}} \leftarrow -\sum_{j=1}^{J} p_{\mathrm{agg}}(\theta^j)\log p_{\mathrm{agg}}(\theta^j)$\;

$H_{\mathrm{agg}}^{\mathrm{norm}} \leftarrow H_{\mathrm{agg}} / \log_2(J)$\;

\KwRet{$(\theta^*, H_{\mathrm{agg}}^{\mathrm{norm}})$}
\end{algorithm}
%%%%%%%%%%%%%%%%%%%%%%%%%%%%%%%%%%%%%%%%%%%%%%%%%%%%%%%%%%%%%%%%%%%%%%

\newpage

\subsection{The Aggregation Baselines}
\label{app:baseline}

%%%%%%%%%%%%%%%%%%%%%%%%%%%%%%%%%%%%%%%%%%%%%%%%%%%%%%%%%%%%%%%%%%%%%%%%%%%%%%%%%%%%%%%%%%%%%%%%
\begin{figure}[h]
    \centering
    \begin{subfigure}[t]{0.48\columnwidth}
        \centering
        \includegraphics[width=\linewidth]{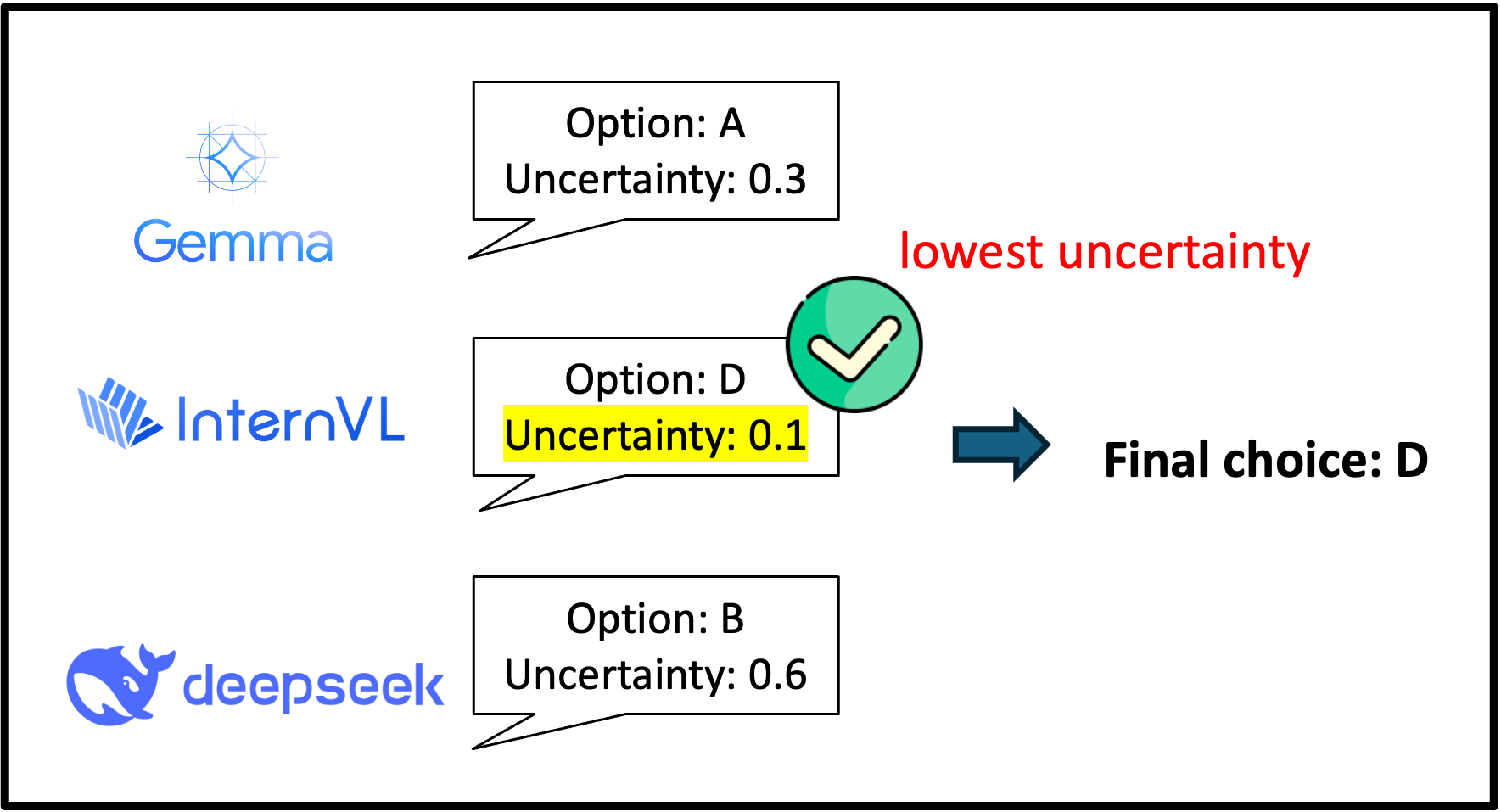}
        \caption{Naive Selection.}
        \label{fig:baseline-ns}
    \end{subfigure}
    \hfill
    \begin{subfigure}[t]{0.48\columnwidth}
        \centering
        \includegraphics[width=\linewidth]{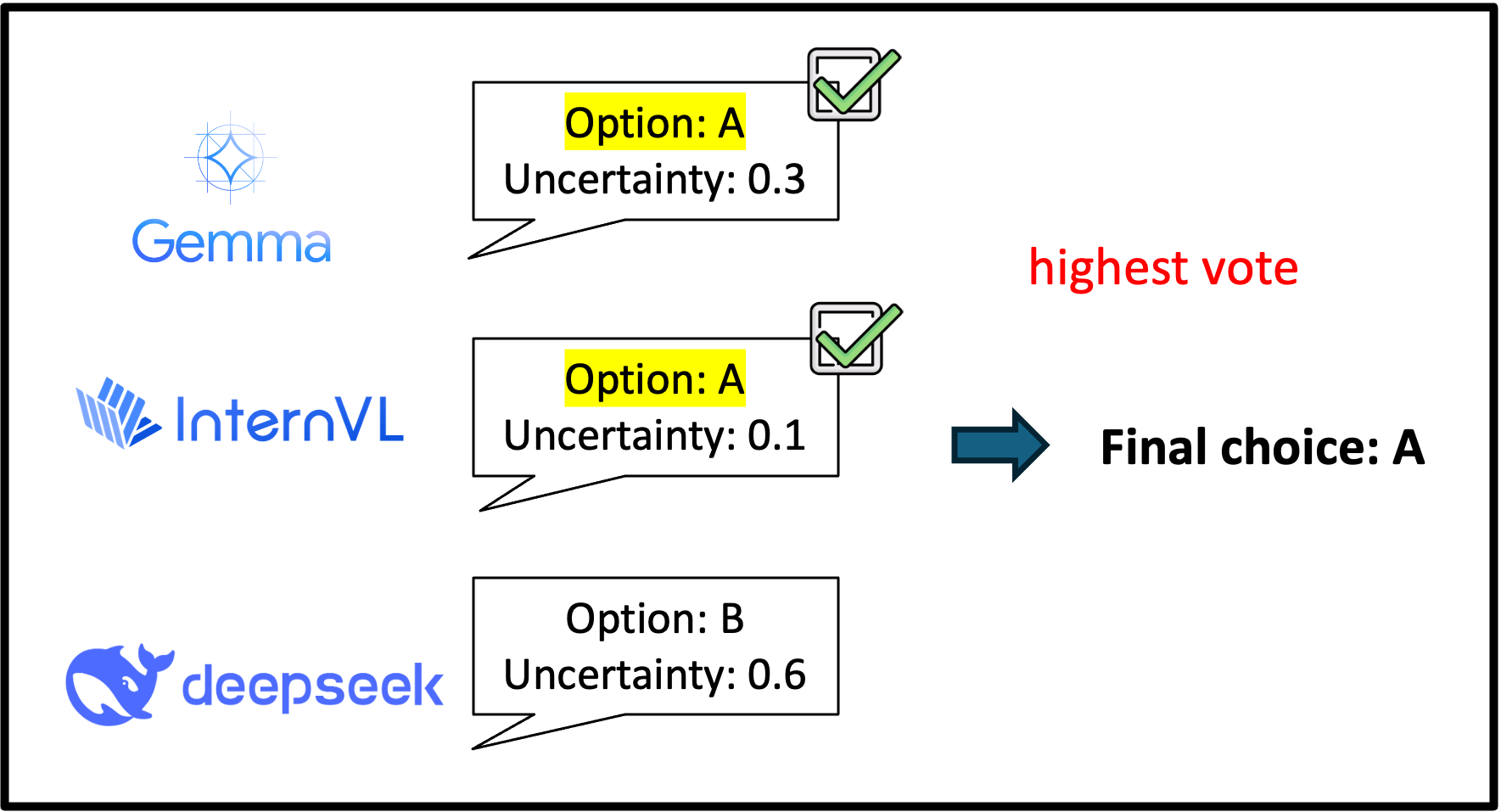}
        \caption{Majority Voting.}
        \label{fig:baseline-mv}
    \end{subfigure}
    \caption{
        \footnotesize{
        \textbf{Illustrations of the aggregation baselines.}
        }
    }
    \label{fig:baseline}
\end{figure}
%%%%%%%%%%%%%%%%%%%%%%%%%%%%%%%%%%%%%%%%%%%%%%%%%%%%%%%%%%%%%%%%%%%%%%%%%%%%%%%%%%%%%%%%%%%%%%%%

Follow the notations defined in Sec.~\ref{sec:method}, we express the math details for the aggregation baselines.
Both baselines start with sampling responses to form a probability vector for each VLM $M_k$, and compute its entropy-based uncertainty by Semantic Entropy \cite{farquhar2024detecting}.
Then the two baselines go into two different routes for VLM systems.

\textbf{Naive Selection.}
This baseline selects a single model purely by its individual entropy-based uncertainty, as illustrated in Fig.~\ref{fig:baseline-ns}.
The model with the lowest entropy is chosen as the system representative: $k^*=\arg \min _k H_k$ .
The system's prediction on the option $\theta^*$ and its uncertainty $H_{\text {NS }}$ are then taken directly from that model:
$$
\theta^*=\arg \max _\theta p_{k^*}(\theta), \quad H_{\text {NS }}=H_{k^{*}}
$$

\textbf{Majority Voting.}
Each model ($M_k$) votes its top-1 probable option $\theta_k^* = \arg\max_{\theta\in\Theta} p_k(\theta)$, where $p_k(\theta)$ is the probability assigned by model $k$ to option $\theta$.
It is illustrated in Fig.~\ref{fig:baseline-mv}.
Across $K$ VLMs, we form the empirical vote distribution:  
$$
p_{\mathrm{MV}}(\theta) = \frac{1}{K}\sum_{k=1}^{K}\mathbf{1}[\theta_k^*=\theta],  
$$
with ($\sum_{\theta\in\Theta_{\mathrm{voted}}}p_{\mathrm{MV}}(\theta)=1$).
The final prediction is then:
$$
\theta^* = \arg\max_{\theta\in\Theta_{\mathrm{voted}}}p_{\mathrm{MV}}(\theta),  
$$
and the system's uncertainty:
$$
H_{\mathrm{MV}} = -\sum_{\theta\in\Theta_{\mathrm{voted}}} p_{\mathrm{MV}}(\theta)\log p_{\mathrm{MV}}(\theta) ,
$$
In case of ties (equal vote counts), it resolves them by selecting the option whose winning model has the higher probability $p_k$.
Note that we normalize the entropy (for both Naive Selection and Majority Vote) by the number of MCQ options $J$, which ensures that the uncertainty measure remains comparable across samples and models, as we did in Sec.~\ref{sec:method}.

% \textbf{Uniform Opinion Pooling (UOP).}
% To isolate the effect of SCOPE's uncertainty weighting, we define an unweighted probabilistic aggregation baseline.
% UOP extends MV into probability space by averaging the aligned probability vectors from all models:

% $$
% p_{\mathrm{UOP}}(\theta)=\frac{1}{K} \sum_{k=1}^K \tilde{p}_k(\theta)
% $$

% The system prediction is then:
% $$
% \theta^*=\arg \max _\theta p_{\mathrm{UOP}}(\theta)
% $$
% and the uncertainty is calculated:
% $$
% H_{\mathrm{UOP}}^{\text {norm }}=-\frac{1}{\log _2 J} \sum_{\theta \in \Theta} p_{\mathrm{UOP}}(\theta) \log _2 p_{\mathrm{UOP}}(\theta)
% $$

% This baseline therefore acts as the unweighted counterpart of SCOPE, revealing the contribution of uncertainty-based weighting.

%%%%%%%%%%%%%%%%%%%%%%%%%%%%%%%%%%%%%%%%%%%%%%%%%%%%%%%%%%%%
%%%%%%%%%%%%%%%%%%%%%%%%%%%%%%%%%%%%%%%%%%%%%%%%%%%%%%%%%%%%
\newpage
\subsection{Additional Experiments Details}
\label{app:exp_detail}

\textbf{Comprehensive VLM List for Experiments. }
Sixteen open-weight, instruction-tuned VLMs are employed, spanning five representative models: \textit{LLaVA-v1.6}, \textit{Gemma-3}, \textit{InternVL3}, \textit{DeepSeek-VL2}, and \textit{Qwen2.5-VL}, and grouped by parameter scale as summarized in Table~\ref{tab:vlm_combo}.
Each model is evaluated using its official inference configuration with identical decoding hyperparameters ($T=1.0$, nucleus sampling $P=0.9$, top-$K$ sampling $K=50$) across all scales to ensure fair and consistent comparison.
Across these models, a total of 52 unique multi-VLM system combinations are constructed for $K\in\{2,3,4,5\}$, covering four parameter scales and five models.

\textbf{Model Settings. }
Following the prior study by \cite{farquhar2024detecting}, we set the temperature $T=1.0$ ($\in [0,1]$), where a higher temperature encourages randomness in generations, with nucleus sampling $P=0.9$ and top-$K$ sampling $K=50$ applied to all VLMs.
In line with the empirical settings of \cite{zhang2024vl}, each model is sampled ten times to capture model's variability for quantifying its uncertainty.
An $\epsilon$ constant of $1\times10^{-6}$ is applied throughout to prevent division by zero when computing entropy and normalization terms.
When a sampled response does not correspond to any valid MCQ option, we assign it a placeholder label of $-1$ and continue.
This preserves the model's original ``opinion" during sampling.
Since such cases may alter the number of effective classes between models, SCoOP dynamically extends each probability vector to maintain a unified dimension across models, which ensures consistent entropy scaling and accurate uncertainty quantification.

\textbf{Computation Details. }
All the experiments are conducted on one NVIDIA Blackwell B200 GPU (180 GB memory) and four Intel Xeon Platinum 8570 CPUs (8 GB RAM per core), ensuring both computational efficiency and reproducibility through standardized resource provisioning.
%%%%%%%%%%%%%%%%%%%%%%%%%%%%%%%%%%%%%%%%%%%%%%%%%%%%%%%%%%%%
\begin{table*}[h]
\centering
\footnotesize
\caption{Comprehensive list of VLMs used in our experiments and the number of possible model combinations per system size $K$.}
\vspace{-5pt}
\label{tab:vlm_combo}
\resizebox{\textwidth}{!}{
\begin{tabular}{l|l|l}
\toprule
\textbf{Model Size} & \textbf{Models} & \textbf{Combination Count} \\
\midrule
Small (2–4B) &
\begin{tabular}[c]{@{}l@{}}\texttt{gemma-3-4b-it}, \texttt{Qwen2.5-VL-3B-Instruct}, \\ \texttt{InternVL3-2B-Instruct}, \texttt{deepseek-vl2-tiny}\end{tabular} &
\begin{tabular}[c]{@{}c@{}}2-VLM: 6 \\ 3-VLM: 4 \\ 4-VLM: 1\end{tabular} \\
\midrule
Medium (12–16B) &
\begin{tabular}[c]{@{}l@{}}\texttt{gemma-3-12b-it}, \texttt{llava-v1.6-vicuna-13b-hf}, \\ \texttt{InternVL3-14B-Instruct}, \texttt{deepseek-vl2-small}\end{tabular} &
\begin{tabular}[c]{@{}c@{}}2-VLM: 6 \\ 3-VLM: 4 \\ 4-VLM: 1\end{tabular} \\
\midrule
Large (27–38B) &
\begin{tabular}[c]{@{}l@{}}\texttt{gemma-3-27b-it}, \texttt{llava-v1.6-34b-hf}, \texttt{Qwen2.5-VL-32B-Instruct}, \\ \texttt{InternVL3-38B-Instruct}, \texttt{deepseek-vl2}\end{tabular} &
\begin{tabular}[c]{@{}c@{}}2-VLM: 10 \\ 3-VLM: 10 \\ 4-VLM: 5 \\ 5-VLM: 1\end{tabular} \\
\midrule
Extra Large (72–78B) &
\begin{tabular}[c]{@{}l@{}}\texttt{llava-next-72b-hf}, \texttt{Qwen2.5-VL-72B-Instruct}, \\ \texttt{InternVL3-78B-Instruct}\end{tabular} &
\begin{tabular}[c]{@{}c@{}}2-VLM: 3 \\ 3-VLM: 1\end{tabular} \\
\bottomrule
\end{tabular}
}
\end{table*}
%%%%%%%%%%%%%%%%%%%%%%%%%%%%%%%%%%%%%%%%%%%%%%%%%%%%%%%%%%%%

%%%%%%%%%%%%%%%%%%%%%%%%%%%%%%%%%%%%%%%%%%%%%%%%%%%%%%%%%%%%
%%%%%%%%%%%%%%%%%%%%%%%%%%%%%%%%%%%%%%%%%%%%%%%%%%%%%%%%%%%%
\newpage
\subsection{Prompt Template}

% \paragraph{Prompt: VLM discrete label generation.}
\label{sec:appendix-prompts}
We apply the following prompt template for all the experiments across three different benchmark datasets, which ensures VLMs generating responses with discrete choices. 
For illustration purposes, in this paper we use (A), (B), (C) to represent choices to avoid confusion with other numbers.
In our code, we use (1), (2), (3) instead for the simplicity of the code.
The number of choices depends on the question, which could be ranged from two to five.

\begin{table}[h]
    \centering
    \caption{Prompt template for VLM generating discrete choices.}
    \vspace{-5pt}
    \begin{tcolorbox}[colframe=black, colback=white, arc=8pt, boxrule=0.8pt] % Rounded corners
    \footnotesize
    \begin{tabular}{p{0.9\textwidth}} % Single-column table inside the rounded box
        $<$image$>$\\
        \\
        \textcolor{blue}{\{benchmark\_question\}}\\
        (A): \textcolor{blue}{\{question\_choice\}} \\
        (B): \textcolor{blue}{\{question\_choice\}} \\
        (C): \textcolor{blue}{\{question\_choice\}} \\
        (D): \textcolor{blue}{\{question\_choice\}} \\
        (E): \textcolor{blue}{\{question\_choice\}} \\
        \\
        This is a single choice question, answer only with choice label in \textcolor{blue}{\{question\_choice\_set\}}.
        % \vspace{3pt}
    \end{tabular}
    \end{tcolorbox} % End rounded box
    \label{tab:template_2}
\end{table}

\end{document}